\pdfoutput=1
\documentclass[conference]{IEEEtran}

\usepackage[T1]{fontenc}
\usepackage[utf8]{inputenc}
\usepackage{cite}
\usepackage{amsmath}
\usepackage{booktabs}
\usepackage{graphicx}
\usepackage{url}
\usepackage[hidelinks]{hyperref}

\newcommand{\orcidlink}[1]{\href{https://orcid.org/#1}{\raisebox{-0.15\height}{\includegraphics[height=1.25em]{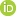}}}}
\newcommand{\rev}[1]{#1}

\graphicspath{{../outputs/figures/png/}{../outputs/massive_secondary/figures/png/}}

\IEEEoverridecommandlockouts
\title{A Cost-Efficient Routing Pipeline for Multilingual Short-Text Classification Using Small Language Models%
\thanks{This paper was accepted for publication at the 16th International Conference on Advanced Computer Information Technologies (ACIT 2026), \url{https://acit.tech/}. \copyright~2026 IEEE. Personal use of this material is permitted. Permission from IEEE must be obtained for all other uses.}}

\author{
\IEEEauthorblockN{Wajdi BEN SAAD}
\IEEEauthorblockA{\textit{Carthago Labs} \\
Paris, France \\
\orcidlink{0000-0002-0939-016X}}
\and
\IEEEauthorblockN{Safa MADIOUNI}
\IEEEauthorblockA{\textit{Universit\'{e} Paris Dauphine-PSL} \\
Paris, France \\
\orcidlink{0009-0004-1025-0887}}
}

\begin{document}

\maketitle

\begin{abstract}
Multilingual short-text classification supports operational systems such as content moderation, customer support routing, and intent recognition, yet aggregate evaluation often hides large differences between high-resource and low-resource languages. Uniform inference policies are simple to deploy, but they assume that all languages are equally well served. In this work, we evaluate a fixed-list routing strategy that keeps stronger languages on a direct multilingual path and selectively sends weaker languages through translation into English before zero-shot classification. The pipeline is fully self-hosted, uses pretrained compact sentence encoders, and requires no task-specific fine-tuning.

We test the approach on two benchmarks chosen to differ in scale and label granularity: a 15-language subset of SIB-200 for seven-way topic classification and a 15-locale subset of MASSIVE for intent classification over an official 60-intent inventory. On SIB-200, the best overall configuration is R1, which translates only the low-resource tier: high-tier and mid-tier Macro-F1 remain unchanged, while low-tier Macro-F1 rises from 0.4632 to 0.6828. On the MASSIVE subset, the same low-tier intervention raises low-tier Macro-F1 from 0.2143 to 0.4417, but the best overall result is obtained by full translation, R3, at Macro-F1 0.4647. Across these two benchmarks, selective translation is a reliable intervention for weaker languages, whereas the optimal routing boundary depends on the task. We therefore report routing through tier-level quality gains and tier-level latency rather than a single global efficiency score.

\end{abstract}


\section{Introduction}
\label{sec:introduction}

Multilingual short-text classification is required in content moderation, search assistance, customer support, and multilingual monitoring systems. A common deployment choice is to apply one inference path to every language, either with a single multilingual encoder or by translating every input into a pivot language. That uniform treatment is convenient, but benchmarks such as XTREME \cite{hu2020xtreme}, SIB-200 \cite{adelani2024sib}, and MASSIVE \cite{fitzgerald2023massive} show broad variation across languages, scripts, and task settings even under the same model family.

Translation-based transfer remains a competitive alternative to direct multilingual inference \cite{unanue2023t3l,artetxe2023revisiting}, but it adds latency, depends on translation-model coverage, and may introduce artifacts \cite{artetxe2020artifacts}. Our work targets a local, cost-aware deployment regime in which compact self-hosted models remain attractive relative to larger generative systems with higher serving overhead \cite{xue2021mt5,lin2022xglm}. The practical question is therefore not only which method is strongest overall, but whether translation should be reserved for the languages that benefit from it most.

In this work, we examine that question with a fixed-list routing policy. We compare multilingual-only inference, translation-only inference, and a static router that sends selected language tiers through a translate-then-classify path. Unlike standard translate-and-test pipelines, our goal is not to replace direct multilingual inference everywhere, but to identify where translation should begin and where it should stop. Both paths use the same prototype-matching rule over English labels, and no task-specific fine-tuning is introduced. We conduct the analysis on a 15-language subset of SIB-200 for topic classification and a 15-locale subset of MASSIVE for intent classification. Although MASSIVE is commonly used for supervised multilingual natural language understanding (NLU), here we use only the public test split for zero-shot inference, with no training or weak supervision.

The main finding is localized rather than universal. On SIB-200, translating only the low-resource tier raises low-tier Macro-F1 from 0.4632 to 0.6828 while the high-tier and mid-tier remain unchanged, and R1 is the strongest overall configuration. On the MASSIVE subset, the same low-tier intervention raises low-tier Macro-F1 from 0.2143 to 0.4417, again without changing the stronger tiers under R1, but the strongest overall result is obtained by full translation, R3. Under this setup, the central design question is not whether routing should be made more complex, but where the translation boundary should be placed for the task at hand.

In this paper, we make three contributions:
\begin{enumerate}
    \item We present a fully self-hosted empirical study of fixed-list routing for multilingual short-text classification without task-specific fine-tuning, evaluated on topic classification and intent classification.
    \item We report the results in a deployment-oriented form based on delta Macro-F1 relative to a multilingual-only baseline and latency measured separately by resource tier.
    \item We show that the routing boundary is dataset-dependent: SIB-200 is best served by translating only the low-tier, whereas the MASSIVE subset reaches its strongest overall result under full translation even though the low-tier R1 gain still persists.
\end{enumerate}

The remainder of this paper is organised as follows: Section~\ref{sec:related_work} reviews prior work on multilingual evaluation, cross-lingual transfer, and routing-related deployment questions. Section~\ref{sec:method} presents our routing pipeline, inference paths, and reporting strategy. Section~\ref{sec:experimental_setup} describes the datasets, language tiers, models, execution environment, and evaluation conditions. Section~\ref{sec:results} reports the main experimental results on SIB-200 and the MASSIVE subset. Section~\ref{sec:discussion} interprets the findings, highlights practical implications, and states the main limitations. Section~\ref{sec:conclusion} concludes the paper and outlines directions for future work.

\section{Related Work}
\label{sec:related_work}

Two strands of prior work are directly relevant here: multilingual evaluation and cross-lingual transfer mechanisms. XTREME \cite{hu2020xtreme} and XTREME-R \cite{asai2021xtremer} established broad multilingual benchmark suites, SIB-200 \cite{adelani2024sib} widened topic-classification evaluation to a much larger set of languages and dialects, and MASSIVE \cite{fitzgerald2023massive} provided a large multilingual NLU resource with a fine-grained intent inventory. Together, these benchmarks show that multilingual performance is not evenly distributed across languages, scripts, or label spaces. That observation motivates our decision to study routing as a language-dependent design choice rather than to assume one universal path.

Our classification pipeline relies on sentence-level representations. Sentence-BERT \cite{reimers2019sentencebert}, LASER \cite{artetxe2019laser}, and LaBSE \cite{feng2022labse} showed that sentence embeddings can support strong zero-shot transfer across many languages. Broader multilingual encoder families such as XLM-R \cite{conneau2020xlmr}, InfoXLM \cite{chi2021infoxlm}, and VECO \cite{luo2021veco} further strengthened the multilingual representation toolkit. This matters here because it lets us compare routing decisions without changing the downstream classifier family. We do not propose a new embedding model; instead, we use compact sentence encoders in a prototype-matching setting so that representation quality remains approximately constant while the routing policy is varied.

Translate-and-test methods provide the second point of comparison. T3L \cite{unanue2023t3l} and the re-evaluation by Artetxe et al. \cite{artetxe2023revisiting} show that translation-based transfer remains competitive. At the same time, translation artifacts can distort downstream behavior \cite{artetxe2020artifacts}, and full translation pipelines introduce non-trivial operational cost. Larger multilingual generative models such as mT5 \cite{xue2021mt5} and XGLM \cite{lin2022xglm} expand multilingual transfer further, but they target a different operating regime from the compact self-hosted setup studied here.

Work on adaptive inference, such as DeeBERT \cite{xin2020deebert}, has shown that selective execution can reduce unnecessary computation when a uniform path is not justified for every input. Our work is related in spirit but differs in granularity: DeeBERT adapts computation depth within a model, whereas we adapt the inference path at the language-tier level under a fixed policy. For local deployment, open translation systems such as OPUS-MT \cite{tiedemann2020opusmt} and broader-coverage models such as NLLB \cite{nllb2024nature} make this kind of routing technically feasible.

The gap we address is therefore specific. Prior work has established strong multilingual encoders, strong translate-and-test pipelines, and adaptive inference within models, but it does not directly ask how a simple fixed language list should place the translation boundary in a fully self-hosted zero-shot classifier. Our work addresses that question by holding the classifier constant, sweeping the routing boundary explicitly, and reporting both quality gains and latency at the resource tier where the intervention is applied.

\section{Method}
\label{sec:method}

\subsection{Task Formulation}

The primary benchmark is seven-way topic classification on SIB-200, and the secondary benchmark is intent classification on a 15-locale subset of MASSIVE. In both cases, let $x$ be an input text and let $\mathcal{Y}$ be the set of English labels for the current dataset. For each encoder, we embed the label texts once to obtain a prototype bank $\{p_y\}_{y \in \mathcal{Y}}$. Given an input representation $h(x)$, prediction is:
\begin{equation}
\hat{y} = \arg\max_{y \in \mathcal{Y}} \cos(h(x), p_y).
\end{equation}

This formulation keeps the decision rule identical across all conditions. The experiment therefore compares inference paths rather than classifier architectures. For SIB-200, the label texts are the seven English topic names. For MASSIVE, the label inventory is extracted programmatically from the official 60-intent metadata, and each intent is represented by a short English description rather than its raw snake-case identifier. This adjustment was adopted after an English-only sanity check showed that the raw identifiers were under-discriminative for prototype matching.

\subsection{Inference Paths}

We evaluate two paths.
\begin{itemize}
    \item \textbf{Multilingual path.} The original text is encoded directly with the multilingual checkpoint \url{sentence-transformers/paraphrase-multilingual-MiniLM-L12-v2}.
    \item \textbf{Translate-then-classify path.} The text is translated locally into English and encoded with the English checkpoint \url{sentence-transformers/paraphrase-MiniLM-L6-v2}.
\end{itemize}

Translation uses Helsinki-NLP OPUS-MT models where a dedicated source-to-English checkpoint exists. If no such checkpoint is available, the system falls back to \url{facebook/nllb-200-distilled-600M}. All translations are cached and reused across routing settings.

\subsection{Routing Policy}

The router is a static lookup from language to inference path. Languages are assigned to three analysis tiers: high, mid, and low. The tiers are not separate models. They are reporting groups used to decide which languages are sent through translation under each routing boundary.

We evaluate four routing configurations.
\begin{itemize}
    \item \textbf{R0:} no translation
    \item \textbf{R1:} translate only the low-tier
    \item \textbf{R2:} translate the mid-tier and low-tier
    \item \textbf{R3:} translate all tiers
\end{itemize}

The multilingual-only baseline is equivalent to R0, and the translation-only baseline is equivalent to R3. We retain the named baselines because they are useful reference points in the tables.

\begin{figure}[t]
    \centering
    \includegraphics[page=1,width=\linewidth]{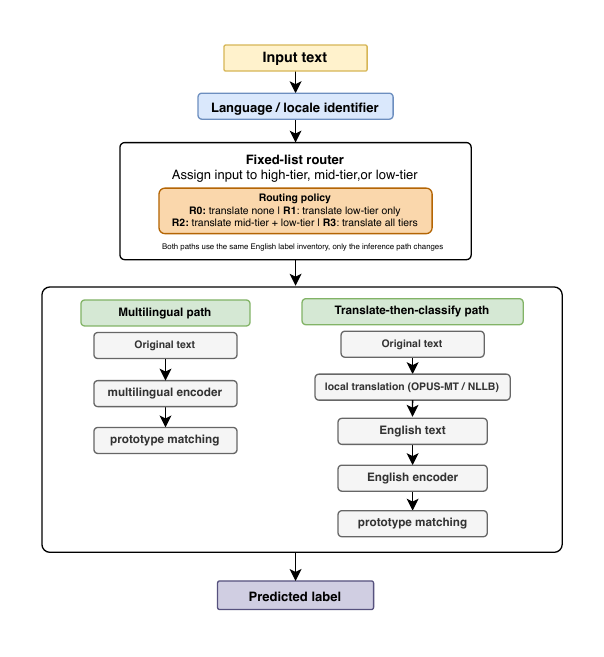}
    \caption{Inference pipeline and routing policy under R0--R3, with each language tier routed to either the multilingual path or the translate-then-classify path.}
    \label{fig:routing_pipeline}
\end{figure}

Figure~\ref{fig:routing_pipeline} summarizes the inference pipeline and the routing policy in one view. The classifier itself remains unchanged across all conditions; only the path taken by a language tier changes.

\subsection{Reporting Strategy}

The headline analysis does not rely on a single composite efficiency ratio. Instead, it reports:
\begin{itemize}
    \item delta Macro-F1 relative to the multilingual-only baseline
    \item mean latency by resource tier
\end{itemize}

This format makes the effect of routing observable at the point where it occurs. If a tier is left unchanged, the corresponding delta is near zero and its latency remains near the multilingual baseline. If a tier is translated, the quality gain and the translation cost can be read together.

\section{Experimental Setup}
\label{sec:experimental_setup}

\subsection{Benchmarks and Language Selection}

We evaluate the routing policy on two benchmarks. The primary benchmark is \texttt{Davlan/sib200} \cite{adelani2024sib}, used here for seven-way topic classification on a 15-language subset with 3{,}060 test examples. The secondary benchmark is the public \texttt{AmazonScience/massive} resource \cite{fitzgerald2023massive}, used here only through its test split for zero-shot intent classification on a 15-locale subset with 44{,}610 test examples. Each selected MASSIVE locale contributes 2{,}974 test examples. The official MASSIVE intent catalog contains 60 intents, although the chosen 15-locale subset observes 59 of them in the selected test examples. The label sets remain in English on both inference paths. For SIB-200, the labels are science/technology, travel, politics, sports, health, entertainment, and geography. For MASSIVE, the official 60-intent inventory is extracted programmatically from the dataset metadata and converted into short English intent descriptions for prototype encoding.

Languages are grouped into high, mid, and low analysis tiers. These tiers are not claimed as a universal resource taxonomy. They are an experimental device used to sweep the translation boundary within each benchmark. The high-tier groups languages that are comparatively well handled by direct multilingual inference, the low-tier groups languages for which translation is more plausible as an intervention, and the mid-tier captures the remaining intermediate cases. The low-tier is aligned across both benchmarks, while the high-tier and mid-tier differ slightly because the available language sets are not identical.

We fixed these groups after inspecting multilingual-only per-language Macro-F1 within each benchmark, but we did not threshold on Macro-F1 alone. Instead, we used the baseline scores to anchor a shared weak-language set across the two benchmarks and then completed the remaining high-tier and mid-tier split with dataset availability and cross-benchmark comparability in mind. The low-tier multilingual-only scores are clearly weak on MASSIVE (Swahili 0.1309, Amharic 0.1501, Telugu 0.1990, Bengali 0.2066, Afrikaans 0.3197) and are again among the weaker languages on SIB-200 (Telugu 0.2780, Bengali 0.3848, Amharic 0.4348, Swahili 0.4537, Afrikaans 0.6911). We therefore treat the tiers as benchmark-specific routing groups rather than as a universal language-resource taxonomy.

\begin{table}[t]
\centering
\caption{Benchmark-specific analysis tiers used in the routing sweep.}
\label{tab:tiers}
\footnotesize
\setlength{\tabcolsep}{3pt}
\begin{tabular}{llp{0.36\linewidth}}
\toprule
Benchmark & Tier & Languages \\
\midrule
SIB-200 & High & English, French, German, Spanish, Chinese \\
\cmidrule(lr){2-3}
 & Mid & Arabic, Turkish, Japanese, Polish, Dutch \\
\cmidrule(lr){2-3}
 & Low & Swahili, Bengali, Telugu, Amharic, Afrikaans \\
\midrule
MASSIVE subset & High & English, French, German, Spanish, Portuguese \\
\cmidrule(lr){2-3}
 & Mid & Arabic, Turkish, Polish, Dutch, Romanian \\
\cmidrule(lr){2-3}
 & Low & Swahili, Telugu, Bengali, Amharic, Afrikaans \\
\bottomrule
\end{tabular}
\end{table}

\subsection{Models}

The multilingual path uses \url{sentence-transformers/paraphrase-multilingual-MiniLM-L12-v2}. The English path uses \url{sentence-transformers/paraphrase-MiniLM-L6-v2}. Both are compact sentence-embedding checkpoints in the Sentence-Transformer family \cite{reimers2019sentencebert}. We chose a compact multilingual encoder rather than a larger checkpoint such as XLM-R \cite{conneau2020xlmr} because the paper targets a self-hosted, cost-aware local deployment setting rather than a model-capacity comparison. Translation uses OPUS-MT \cite{tiedemann2020opusmt} when a language-specific source-to-English checkpoint exists and NLLB \cite{nllb2024nature} otherwise. No task-specific fine-tuning is performed.

\subsection{Conditions}

We evaluate three named conditions:
\begin{itemize}
    \item \textbf{Multilingual-only:} every language uses the multilingual path
    \item \textbf{Translation-only:} every language uses translate-then-classify
    \item \textbf{Routing:} the fixed-list router applies one of R0 to R3
\end{itemize}

Within SIB-200, Bengali and Afrikaans use OPUS-MT, whereas Amharic, Swahili, and Telugu use the NLLB fallback. Within the MASSIVE subset, French, German, Spanish, Arabic, Turkish, Polish, Dutch, Bengali, and Afrikaans use OPUS-MT, whereas Portuguese, Romanian, Swahili, Telugu, and Amharic use the NLLB fallback. This mixed backend is part of the evaluated system rather than a post hoc adjustment.

\subsection{Execution Environment and Metrics}

The reported runs were executed on the same local workstation with an Intel Core i7-9750H processor, 16\,GB RAM, Python 3.11.14, PyTorch 2.2.2, Transformers 4.57.6, and Datasets 4.8.4. The purpose of this setup is not to maximize throughput, but to keep the measurements tied to a reproducible local deployment scenario. For the MASSIVE subset, NLLB-backed translation locales are executed locally with single-beam decoding so the translation stage remains stable and fully cached.

For each condition, we report overall Macro-F1, accuracy, and mean latency. Our tables use the mean of three cached reruns, with standard deviations retained for audit. The main analysis reports:
\begin{itemize}
    \item delta Macro-F1 against the multilingual-only baseline
    \item mean latency for each resource tier
\end{itemize}

For repeated-use settings with cached translations, effective latency can also be summarized as $t_{\mathrm{classify}} + t_{\mathrm{translate}} / N$, where $t_{\mathrm{classify}}$ is the observed mean classification latency, $t_{\mathrm{translate}}$ is the observed mean translation latency for the relevant tier, and $N$ is the average number of reuses per unique input. We also retain the full threshold sweep from R0 to R3, because the sweep is used to identify where translation should stop. On SIB-200, that sweep reveals a localized low-tier benefit. On the MASSIVE subset, it shows that the same low-tier gain persists but that the strongest overall boundary is broader.

The implementation, configuration files, notebooks, and experiment artifacts will be made available at \url{https://github.com/WajdiBenSaad/multilingual-routing-classifier}.

\section{Results}
\label{sec:results}

Unless otherwise noted, all reported values are means over three cached reruns.

\subsection{Primary Benchmark: SIB-200}

Table~\ref{tab:overall} summarizes the SIB-200 routing sweep as means over three cached reruns. The multilingual-only baseline and R0 are numerically identical, as expected. The strongest overall result is obtained by R1, which translates only the low-tier. Moving the boundary to R2 lowers Macro-F1 from 0.7403 to 0.7195 while increasing mean latency from 0.27534\,s to 0.40952\,s. Full translation, represented by R3 and the translation-only baseline, remains below R1 on overall quality and is slower still.

\begin{table}[t]
\centering
\caption{Overall routing sweep on the SIB-200 subset.}
\label{tab:overall}
\begin{tabular}{lccc}
\toprule
Condition & Macro-F1 & Accuracy & Latency (s) \\
\midrule
Multilingual-only & 0.6687 & 0.6614 & 0.0036 \\
R0 & 0.6687 & 0.6614 & 0.0036 \\
R1 & \textbf{0.7403} & \textbf{0.7415} & 0.2753 \\
R2 & 0.7195 & 0.7193 & 0.4095 \\
R3 & 0.6979 & 0.6961 & 0.5380 \\
Translation-only & 0.6979 & 0.6961 & 0.5380 \\
\bottomrule
\end{tabular}
\end{table}

The central comparison is R1 against multilingual-only. Table~\ref{tab:tierdelta} shows that the high-tier and mid-tier are unchanged to four decimal places, whereas the low-tier improves by +0.2196 Macro-F1. The latency increase is likewise localized: high-tier and mid-tier latency remain close to the multilingual baseline, while low-tier mean latency rises to 0.8188\,s because translation is invoked only for that tier.

\begin{table}[t]
\centering
\caption{SIB-200: R1 versus multilingual-only by resource tier.}
\label{tab:tierdelta}
\begin{tabular}{lccccc}
\toprule
Tier & Base F1 & R1 F1 & $\Delta$F1 & Base Lat. & R1 Lat. \\
\midrule
High & 0.7700 & 0.7700 & 0.0000 & 0.00345 & 0.00350 \\
Mid & 0.7674 & 0.7674 & 0.0000 & 0.00358 & 0.00369 \\
Low & 0.4632 & \textbf{0.6828} & \textbf{+0.2196} & 0.00385 & 0.81882 \\
\bottomrule
\end{tabular}
\end{table}

Across the three cached reruns, Macro-F1 and accuracy were identical at the saved precision for every SIB-200 condition. Standard deviations are therefore informative mainly for latency, where the observed variation remains at the fourth decimal place or smaller. We also tested a reporting sensitivity case in which Arabic and Japanese are moved from the mid-tier to the high-tier. That regrouping leaves the low-tier R1 gain unchanged at +0.2196 Macro-F1 and leaves overall Macro-F1 unchanged at 0.7403. Within the low-tier, the largest SIB-200 gains occur in Telugu (+0.4105), Bengali (+0.2914), Amharic (+0.2549), and Swahili (+0.2011), whereas Afrikaans improves only marginally (+0.0135).

\subsection{Secondary Benchmark: MASSIVE Subset}

Table~\ref{tab:secondary_overall} summarizes the MASSIVE routing sweep as means over three cached reruns. The overall pattern differs from SIB-200 in one important respect. The localized low-tier intervention under R1 is again strong, but the best overall configuration is no longer selective routing. R1 improves overall Macro-F1 from 0.3785 to 0.4434, R2 reaches 0.4589, and the strongest overall result is obtained by full translation, R3, with Macro-F1 0.4647 and accuracy 0.4943. At the reported precision, R3 and the translation-only baseline are numerically tied.

\begin{table}[t]
\centering
\caption{Overall routing sweep on the MASSIVE subset.}
\label{tab:secondary_overall}
\begin{tabular}{lccc}
\toprule
Condition & Macro-F1 & Accuracy & Latency (s) \\
\midrule
Multilingual-only & 0.3785 & 0.3855 & 0.0022 \\
R0 & 0.3785 & 0.3855 & 0.0022 \\
R1 & 0.4434 & 0.4704 & 0.0620 \\
R2 & 0.4589 & 0.4856 & 0.1139 \\
R3 & \textbf{0.4647} & \textbf{0.4943} & 0.1515 \\
Translation-only & \textbf{0.4647} & \textbf{0.4943} & 0.1515 \\
\bottomrule
\end{tabular}
\end{table}

The localized pattern under R1 nevertheless persists. Table~\ref{tab:secondary_tierdelta} shows that high-tier and mid-tier Macro-F1 remain unchanged to four decimal places, whereas low-tier Macro-F1 rises from 0.2143 to 0.4417, a gain of +0.2275, with low-tier mean latency rising from 0.0023\,s to 0.1817\,s.

\begin{table}[t]
\centering
\caption{MASSIVE: R1 versus multilingual-only by resource tier.}
\label{tab:secondary_tierdelta}
\begin{tabular}{lccccc}
\toprule
Tier & Base F1 & R1 F1 & $\Delta$F1 & Base Lat. & R1 Lat. \\
\midrule
High & 0.4645 & 0.4645 & 0.0000 & 0.00185 & 0.00198 \\
Mid & 0.4199 & 0.4199 & 0.0000 & 0.00182 & 0.00196 \\
Low & 0.2143 & \textbf{0.4417} & \textbf{+0.2275} & 0.00185 & 0.18170 \\
\bottomrule
\end{tabular}
\end{table}

The per-language gains inside the low-tier are concentrated in Swahili (+0.3116), Telugu (+0.2796), Amharic (+0.2406), Bengali (+0.2297), and Afrikaans (+0.1375). Table~\ref{tab:lowtier_deltas} places these low-tier deltas side by side with the SIB-200 values. What changes relative to SIB-200 is not the existence of the low-tier effect, but the best global boundary: on MASSIVE, extending translation to the mid-tier and then to all tiers continues to improve overall performance.

\subsection{Cross-Dataset Comparison}

The most compact cross-dataset summary is the best boundary selected by each benchmark. SIB-200 reaches its strongest result under R1 with Macro-F1 0.7403 and an R1 low-tier gain of +0.2196. The MASSIVE subset reaches its strongest result under R3 with Macro-F1 0.4647, while its R1 low-tier gain is +0.2275. The repeated low-tier gain therefore generalizes across benchmarks, whereas the best global boundary does not.

\begin{table}[t]
\centering
\caption{Best observed configuration by benchmark.}
\label{tab:crossbest}
\footnotesize
\setlength{\tabcolsep}{3pt}
\begin{tabular}{lcccc}
\toprule
Dataset & Best cfg. & Macro-F1 & Accuracy & R1 low $\Delta$F1 \\
\midrule
SIB-200 & R1 & 0.7403 & 0.7415 & +0.2196 \\
MASSIVE & R3 & 0.4647 & 0.4943 & +0.2275 \\
\bottomrule
\end{tabular}
\end{table}

\begin{table}[t]
\centering
\caption{Low-tier per-language R1 deltas versus multilingual-only.}
\label{tab:lowtier_deltas}
\footnotesize
\setlength{\tabcolsep}{4pt}
\begin{tabular}{lccccc}
\toprule
Dataset & Swahili & Telugu & Bengali & Amharic & Afrikaans \\
\midrule
SIB-200 & +0.2011 & +0.4105 & +0.2914 & +0.2549 & +0.0135 \\
MASSIVE & +0.3116 & +0.2796 & +0.2297 & +0.2406 & +0.1375 \\
\bottomrule
\end{tabular}
\end{table}

The low-tier is served by a mixed local translation backend in both benchmarks. On SIB-200, Bengali and Afrikaans use OPUS-MT, whereas Telugu, Amharic, and Swahili use the NLLB fallback. On the MASSIVE subset, Bengali and Afrikaans again use OPUS-MT, whereas Swahili, Telugu, and Amharic use the NLLB fallback; Portuguese and Romanian also use NLLB outside the low-tier. Gains are therefore not confined to one backend family.

Table~\ref{tab:cross_tier_compact} isolates the most stable cross-dataset pattern in compact form. In both benchmarks, high-tier and mid-tier deltas remain exactly zero at the reported precision under R1, whereas the low-tier carries the entire quality gain and nearly all added latency.

\rev{Under R1, the high-tier and mid-tier examples remain on the multilingual path by construction. Their predictions are therefore identical to the multilingual-only baseline, with zero fixed examples and zero regressed examples in both benchmarks. The paired analyses below focus on the translated low-tier, where predictions actually change.}

\begin{table}[t]
\centering
\caption{Cross-dataset R1 summary by tier. Delta Macro-F1 is measured against multilingual-only.}
\label{tab:cross_tier_compact}
\footnotesize
\setlength{\tabcolsep}{4pt}
\begin{tabular}{llcc}
\toprule
Dataset & Tier & $\Delta$F1 & R1 Latency (s) \\
\midrule
SIB-200 & High & 0.0000 & 0.00350 \\
\cmidrule(lr){2-4}
 & Mid & 0.0000 & 0.00369 \\
\cmidrule(lr){2-4}
 & Low & +0.2196 & 0.81882 \\
\midrule
MASSIVE & High & 0.0000 & 0.00198 \\
\cmidrule(lr){2-4}
 & Mid & 0.0000 & 0.00196 \\
\cmidrule(lr){2-4}
 & Low & +0.2275 & 0.18170 \\
\bottomrule
\end{tabular}
\end{table}

\begin{table*}[t]
\centering
\begingroup
\caption{Qualitative paired error analysis for R1 versus multilingual-only. ``Fixed'' counts examples that are wrong under multilingual-only and correct under R1; ``regressed'' counts the reverse.}
\label{tab:paired_error_analysis}
\footnotesize
\setlength{\tabcolsep}{6pt}
\begin{tabular}{llrrrr}
\toprule
Dataset & Scope & Base Correct & R1 Correct & Fixed by R1 & Regressed by R1 \\
\midrule
SIB-200 & Low-tier & 450 & 695 & 329 & 84 \\
SIB-200 & Afrikaans & 142 & 144 & 23 & 21 \\
MASSIVE & Low-tier & 3233 & 7019 & 4606 & 820 \\
MASSIVE & Afrikaans & 1016 & 1430 & 658 & 244 \\
\bottomrule
\end{tabular}
\endgroup
\end{table*}

\rev{Table~\ref{tab:paired_error_analysis} shows that the R1 low-tier gains reflect many more corrected predictions than newly introduced errors. The Afrikaans rows explain why its net gain is smaller: on SIB-200, translation fixes 23 Afrikaans examples but regresses 21, leaving only a small net change.}

\begin{table*}[t]
\centering
\begingroup
\caption{Paired significance check for the translated low-tier under R1. The test is McNemar-style over saved paired example-level predictions, not over the deterministic cached reruns.}
\label{tab:paired_significance}
\footnotesize
\setlength{\tabcolsep}{5pt}
\begin{tabular}{lllrrrr}
\toprule
Dataset & Comparison & Scope & $\Delta$ Macro-F1 & Fixed & Regressed & Paired test \\
\midrule
SIB-200 & R1 vs multilingual-only & Low-tier & +0.2196 & 329 & 84 & $p < 0.001$ \\
MASSIVE & R1 vs multilingual-only & Low-tier & +0.2275 & 4606 & 820 & $p < 0.001$ \\
\bottomrule
\end{tabular}
\endgroup
\end{table*}

\rev{Table~\ref{tab:paired_significance} indicates that the main low-tier R1 improvements are not explained by a small number of isolated prediction changes. The repeated cached reruns are retained as a repeatability check, while the paired test uses example-level correctness changes between multilingual-only and R1.}

Across both benchmarks, the rerun standard deviations are effectively zero at the stored precision for Macro-F1 and accuracy because the evaluation is deterministic once prototypes and cached translations are fixed. The reruns are therefore most useful as a repeatability audit and as a check that the latency measurements remain stable.

Because translations are cached, the measured per-example latency also admits a repeated-use interpretation. We do not treat effective cached latency as a replacement for the measured end-to-end latency, but it remains relevant for deployment scenarios in which the same or highly similar queries recur.

\section{Discussion}
\label{sec:discussion}

Our results support a practical but limited claim. Fixed-list routing is not a universal replacement for multilingual inference; it is a boundary-selection problem whose answer depends on the benchmark. On SIB-200, the high-tier and mid-tier are already well handled by the multilingual path, whereas the low-tier benefits substantially from translation into English. On the MASSIVE subset, that same low-tier rescue effect remains strong, but the best overall setting extends translation further.

That distinction matters for system design. A simple lookup table was sufficient to expose the main structural effect, but the effect was not identical across tasks. On SIB-200, once translation crosses the low-tier boundary and is extended to the mid-tier, the gains do not persist. This is consistent with the multilingual encoder already providing adequate coverage for mid-tier languages, making translation more likely to add noise than useful transfer for that benchmark. On the MASSIVE subset, the opposite trend appears: the low-tier benefit under R1 is retained, but broader translation continues to improve the overall result. Our current explanation is a hypothesis rather than a demonstrated causal finding. MASSIVE uses a denser 60-intent inventory, and even in the selected subset the evaluation spans 59 observed intents, which may make the direct multilingual prototype path more sensitive to fine-grained lexical distinctions than the seven-way topic setup in SIB-200.

\rev{Afrikaans is an informative outlier within the low-tier. On SIB-200, its multilingual-only Macro-F1 is already 0.6911 and R1 raises it only to 0.7045, a gain of +0.0135; the paired error analysis shows 23 fixed Afrikaans examples and 21 regressed examples. On MASSIVE, Afrikaans again has the strongest low-tier direct baseline, rising from 0.3197 to 0.4571 under R1, but it remains the smallest low-tier gain (+0.1375), with 658 fixed examples and 244 regressed examples. This pattern is consistent with prior cross-lingual NLP evidence that embedding similarity can predict transfer success \cite{idris2026embedding}, but we treat the proximity explanation as a hypothesis rather than a causal finding.}

Our study also illustrates a reproducibility point that is often omitted in benchmark reports. A fully local multilingual pipeline may require a hybrid translation backend because bilingual model availability is uneven. Our final system combines OPUS-MT and NLLB for exactly that reason.

For practical deployment, a team can begin with direct multilingual inference for all languages, identify the subset that remains weak under that baseline, and reserve translation only for that subset. Our results suggest that such a policy does not need a complicated learned router to be useful.

The two-benchmark design adds a methodological lesson. SIB-200 and MASSIVE do not simply differ in scale; they probe different classification regimes. Evaluating both lets us separate what appears stable across tasks, namely the repeated low-tier gain, from what remains benchmark-specific, namely the location of the strongest global routing boundary.

Several limitations should be stated plainly. First, two datasets and two task families are still not enough to treat any discovered boundary as universal. Second, the tier assignment is hand-crafted for the experiment; it is not learned from data. Third, the MASSIVE benchmark requires a reported design choice: short English intent descriptions are used as prototype texts because the raw snake-case intent names were too weak in the English-only sanity check. Fourth, the selected MASSIVE subset observes 59 of the 60 official intents, even though the full official inventory is retained in the prototype bank. Fifth, the measurements are local and model-specific, which suits our deployment framing but does not cover every production environment.

Finally, the latency numbers require interpretation rather than compression into a single scalar. Translation cost is substantial in per-example terms, yet it is localized to the translated tier and can be amortized in repeated-use settings through caching. For this reason, we report quality deltas and tier-specific latency side by side.

\section{Conclusion}
\label{sec:conclusion}

In this work, we evaluated a fully self-hosted and cost-aware routing pipeline for multilingual short-text classification on two benchmarks. Using pretrained compact sentence encoders and no task-specific fine-tuning, we compared multilingual-only inference, translation-only inference, and a fixed routing sweep over three resource tiers.

The main result is that the best routing boundary is not universal. On SIB-200, translating only the low-resource tier substantially improves low-tier Macro-F1 while leaving the stronger tiers unchanged, and that selective policy is the best overall configuration. On the MASSIVE subset, the same low-tier intervention again yields a large improvement, but the strongest overall result is obtained by full translation, R3. Under the present setup, the routing boundary is therefore the central practical design choice.

\rev{The main quantitative outcomes are as follows. On SIB-200, R1 is the best configuration, reaching Macro-F1 0.7403 and improving low-tier Macro-F1 by +0.2196 over multilingual-only. On the MASSIVE subset, R3 is the best overall configuration with Macro-F1 0.4647, while the R1 low-tier intervention still improves low-tier Macro-F1 by +0.2275.}

Two aspects of our work are worth emphasizing. First, the evidence supports reporting routing effects at the tier level rather than through a single global efficiency score. That makes it possible to see where translation helps, where it does not, and how much latency is actually incurred by the affected languages. Second, the repeated low-tier gain across both benchmarks is the most stable empirical result we observe. On SIB-200, low-tier R1 Macro-F1 rises by +0.2196. On the MASSIVE subset, low-tier R1 Macro-F1 rises by +0.2275. What changes across datasets is not whether the low-tier benefits, but how far translation should be extended beyond it.

The contribution is therefore empirical and operational. We show where translation helps, where it does not, and how that decision can be implemented with a simple fixed-list policy using only local resources. More broadly, our results argue for reporting multilingual deployment outcomes at the point where interventions are applied rather than collapsing all languages into a single system-level score. In our experiments, that perspective reveals two distinct outcomes: SIB-200 favors targeted low-tier translation, whereas the larger and harder MASSIVE benchmark favors broader translation. That difference is not a weakness. It is the main cross-dataset finding.

Future work can extend the same analysis to additional datasets, replace the hand-crafted boundary with a learned policy, and test whether task-specific fine-tuning changes the boundary location. It would also be useful to study whether finer-grained language clustering or confidence-aware fallback rules can recover some of the MASSIVE gains without extending translation to every tier. Even without those additions, this work already establishes a useful baseline for deployment-oriented multilingual evaluation: test the routing boundary directly, report the gain and the cost at the tier where the intervention is applied, document any hybrid translation backend honestly, and do not assume that the same boundary will transfer unchanged across tasks.

\bibliographystyle{IEEEtran}
\bibliography{references}

\end{document}